\documentclass{article}
\usepackage{amsmath}
\usepackage{graphicx}
\usepackage{bm}

\usepackage[final]{nips_2016}

\usepackage{array}

\usepackage[utf8]{inputenc} % allow utf-8 input
\usepackage[T1]{fontenc}    % use 8-bit T1 fonts
\usepackage{hyperref}       % hyperlinks
\usepackage{url}            % simple URL typesetting
\usepackage{booktabs}       % professional-quality tables
\usepackage{amsfonts}       % blackboard math symbols
\usepackage{nicefrac}       % compact symbols for 1/2, etc.
\usepackage{microtype}      % microtypography

\usepackage{tikz}

\usepackage{cancel}
\usepackage{xargs}
 \usepackage[colorinlistoftodos,prependcaption,textsize=tiny]{todonotes}
\newcommandx{\unsure}[2][1=]{\todo[linecolor=green,backgroundcolor=green!25,bordercolor=green,#1]{#2}}
\newcommandx{\change}[2][1=]{\todo[linecolor=blue,backgroundcolor=blue!25,bordercolor=blue,#1]{#2}}
\newcommandx{\improvement}[2][1=]{\todo[linecolor=red,backgroundcolor=red!25,bordercolor=red,#1]{#2}}

\newcommand\blfootnote[1]{%
  \begingroup
  \renewcommand\thefootnote{}\footnote{#1}%
  \addtocounter{footnote}{-1}%
  \endgroup
}

\newcommand*\samethanks[1][\value{footnote}]{\footnotemark[#1]}

\author{Nikolay Jetchev\thanks{These authors contributed equally to this work.} \And Urs Bergmann\samethanks \And Roland Vollgraf\\
Zalando Research\\ \{nikolay.jetchev,urs.bergmann,roland.vollgraf\}@zalando.de\\}
\usepackage[tight]{subfigure}

\newtheorem{theorem}{Proposition}

\title{Texture Synthesis with Spatial Generative Adversarial Networks}
\begin{document}
\maketitle
%CoCoGAN: Conv to Conv Generative Adversarial Network
\section{Abstract}
% note: removed the following Absatz, because details on GANs shouldn't go in abstract
%Generative adversarial networks (GANs)~\cite{Goodfellow14} are a recent approach to training generative models of data by learning simultaneously two models $G$ and $D$. A generative model $G$ is learned that can capture the data distribution and sample from it  given an input noise vector. Its adversary is a discriminative model $D$ that estimates how close a sample is to the true distribution and learns to distinguish between the real data and the samples of $G$.
Generative adversarial networks (GANs)~\cite{Goodfellow14}  are a recent approach to train generative models of data, which have been shown to work particularly well on image data.
In the current paper we introduce a new model for texture synthesis based on GAN learning. By extending the input noise distribution space from a single vector to a whole spatial tensor, we create an architecture with properties well suited to the task of texture synthesis, which we call spatial GAN (SGAN). 
%We demonstrate the power of networks created in an adversarial fashion to create high quality natural looking textures. 
To our knowledge, this is the first successful completely data-driven texture synthesis method based on GANs.

Our method has the following features which make it a state of the art algorithm for texture synthesis:
high image quality of the generated textures, very high scalability w.r.t. the output texture size, fast real-time forward generation, the ability to fuse multiple diverse source images in complex textures. To illustrate these capabilities we present multiple experiments with different classes of texture images and use cases. We also discuss some limitations of our method with respect to the types of texture images it can synthesize, and compare it to other neural techniques for texture generation.

\section{Introduction}
%The paper will proceed by giving an overview of classical and more recent approaches to texture synthesis.
%TODO already here a motivating picture ?
%The problem of texture synthesis can be formulated as follows:  given a small sample of
%a "texture"  generate a larger similar-looking image 
\subsection{Background: texture synthesis}
A texture can be defined as an image containing repeating patterns with some amount of randomness. More formally, a texture is a realization of a stationary ergodic stochastic process~\cite{DCC2013}.\blfootnote{Our source code is available at \url{https://github.com/zalandoresearch/spatial_gan}}
%, see~\cite{LLH04} for a more detailed categorization.% of textures accoring to their randomness.
The goal of visual texture analysis is to infer a generating process from an example texture, which then allows to generate arbitrarily many new samples of that texture - hence performing texture synthesis.
Success in that task is judged primarily by visual quality and closeness to the original texture as estimated by human observers, but also by other criteria which may be application specific~\cite{egst.20091063}, e.g. the speed of analysis and synthesis, ability to generate diverse textures of arbitrary size, the ability to create smoothly morphing textures. %TODO mention any computer graphics applications, why is texture synthesis important? 

Approaches to do that fall in two broad categories. %\improvement{Maybe 3 categories: non-parametric, optimizing for some descriptor, and directly modeling the data distribution?}.
Non-parametric  techniques resample pixels~\cite{EfrosP} or whole patches~\cite{EfrosQ} from example textures, effectively randomizing the input texture in ways that preserve its visual perceptual properties. They can produce high quality images, but have two drawbacks: (i) they do not "learn" any models of the textures of interest but just reorder the input texture using local similarity, and (ii) they can be time-consuming when large textures should be synthesized because of all the search routines involved. There are methods to accelerate example-based techniques~\cite{egst.20091063}, but this requires complicated algorithms.

\begin{figure}
\centering
\subfigure[Input]{ \begin{tikzpicture}
    \node[anchor=south west,inner sep=0] at (0,0) {\includegraphics[height=4.5cm]{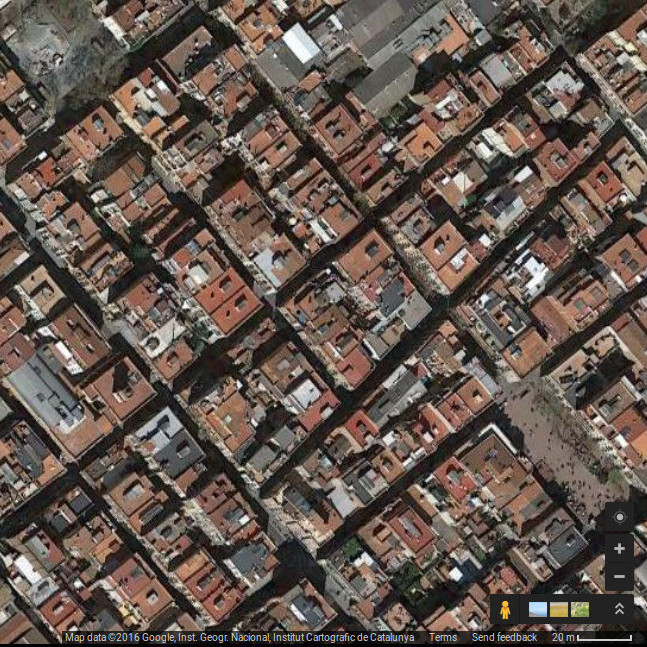}};
    \draw[yellow,dotted,ultra thick] (0,4.5) rectangle (0.87,4.5-0.87);
\end{tikzpicture} }
\subfigure[SGAN5]{\includegraphics[height=4.5cm]{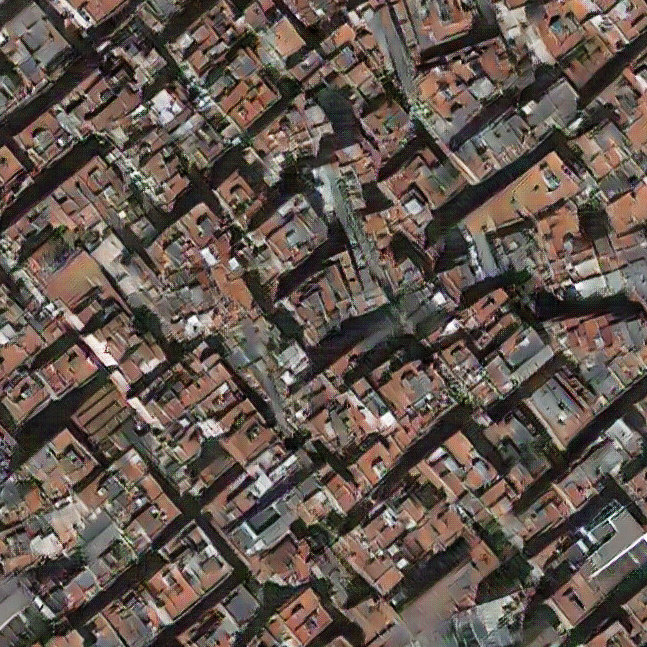} }
\subfigure[Gatys et al.]{\includegraphics[height=4.5cm]{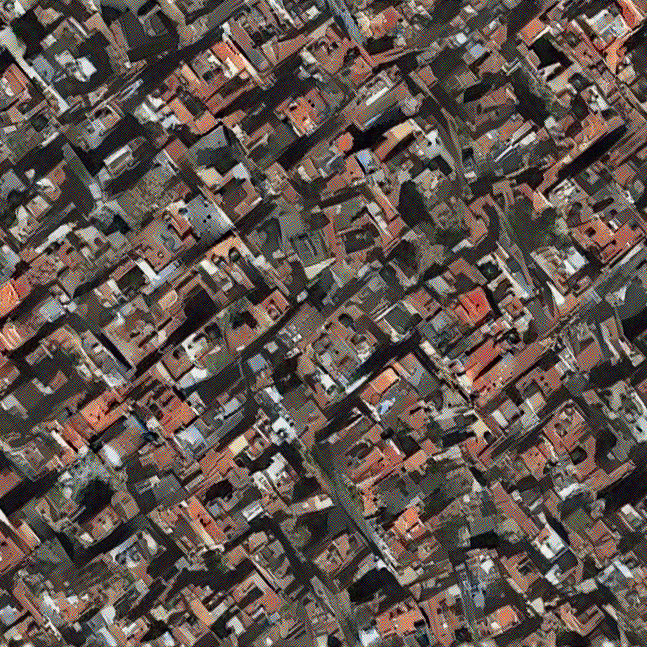}} 

\caption{Learning a texture from a satellite image of Barcelona  of size 1371x647 pixels. We visualize a 647x647 pixel subset of the training and generated images, and for comparison draw the 125 pixel SGAN5 receptive field (yellow box, top-left corner of the left image). Our adversarial approach to texture synthesis SGAN (with 5 layers) generates a city texture of higher visual quality (e.g. clearly visible city streets) than the output of the method of Gatys~\cite{Gatys2015b}. 
%\( SE\cancel{M} \)
\label{imgbarca}}
\end{figure}

The second category of texture synthesis methods is based on matching statistical properties or descriptors of images. 
Texture synthesis is then equivalent to finding an image with similar descriptors, usually by solving an optimization problem in the space of image pixels. 
% The second category of texture synthesis methods is based on parametric models of "good" looking textures. Usually a statistical property of an image is defined and used as a descriptor of the target texture. Any image with descriptor "close" enough to the target texture is considered good. Thus texture synthesis is equivalent to finding an image with the "right" statistics according to the parametric model, usually by solving an optimization problem in the space of image pixels. %Randomization is achieved by starting the optimization process with a different random seed.
The work of Portilla and Simoncelli~\cite{Portilla:2000} is a notable example of this approach, which yields very good image quality for some textures. Carefully designed descriptors over spatial locations, orientations, and scales are used to represent statistics over target textures. 
% redudant to the more abstract version I'd say: Synthesis is done by finding images (by constraint projection) with the same statistics as the target texture.

%\subsection{Deep learning approaches to texture synthesis}
Gatys et al.~\cite{Gatys2015b} present a more data driven parametric approach to allow generation of high quality textures over a variety of natural images. Using filter correlations in different layers of the  convolutional networks -- trained discriminatively on large natural image collections -- results in a powerful technique that nicely captures expressive image statistics. However, creating a single output texture requires solving an optimization problem with iterative backpropagation, which is costly -- in time and memory. % -- for a large output texture size.
%However, speed can be an issue with this method since creating a single output texture requires solving an optimization problem with iterative backpropagation in order to find an image with the desired statistics. This can be quite costly -- in terms of both computational speed and memory -- for a large output texture size.

%his work utilizes the power of convolutional neural networks --  -- to represent rich image statistics and

Recent papers~\cite{ulyanov16texture,Johnson2016Perceptual} deal with that problem and train feed-forward convolutional networks in order to speed up the texture synthesis approach of~\cite{Gatys2015b}. Instead of doing the costly optimization of the output image pixels, they utilize powerful deep learning networks that are trained to produce images minimizing the loss. A separate network is trained for each texture of interest and can then quickly create an image with the desired statistics in one forward pass. 

%There are also methods for generative learning of parametric texture models. 

A generative approach to texture synthesis~\cite{Theis2015c} uses a recurrent neural network to learn the pixel probabilities and statistical dependencies of natural images. They obtain good texture quality on many image types, but their method is computationally expensive and this makes it less practical for texture generation in cases where size and speed matter.
%LSTM methods also interesting, but usually more difficult to scale up

%, and images are created by minimizing a loss function.

%the work of Gatys for texture

%Haager for pyramid based texture analysis

\subsection{An adversarial approach to texture synthesis}
We will present a novel class of generative parametric models for texture synthesis, using a fully convolutional architecture trained employing an adversarial criterion.

As introduced in~\cite{Goodfellow14}, GANs train a generative model $G$ that captures the data distribution and a discriminator $D$ that attempts to separate generated from training data.
%classify whether or not samples are from the true training data distribution.
Radford et al.~\cite{RadfordMC15} improved the GAN architecture by using deep convolutional layers with (fractional) stride~\cite{DB15a} and batch normalization~\cite{IoffeS15}. %, calling this method DCGAN.
Overall, GANs are powerful enough to generate natural looking images of high quality (but low pixel resolution) that can confuse even human observers~\cite{lapgan}.%, and our novel texture synthesis approach would 

However, in GANs the size of the output image (e.g. 64x64 pixels) is hard coded into the network architecture. This is a limitation for texture synthesis, where much larger textures and multiple sizes may be required.
Laplacian pyramids have been used to generate images of increasing size~\cite{lapgan}, gradually adding more details to the images with stacked conditional GANs. However, that technique is still limited in the output image sizes it can handle because it needs to train GAN models with increasing complexity for each scale in the image pyramid. The scale levels must also be specified in advance, so the method cannot create output of arbitrary size.

%Our work can be seen as an extension of the architectures of DCGAN and a novel application of GAN methods.
In our work we will input images of textures (possibly of high pixel resolution) as the data distribution that the GAN must learn. However, we will modify the DCGAN architecture~\cite{RadfordMC15} to allow for scalability and the ability to create any desired output texture size by employing a purely convolutional architecture without any fully connected layers. The SGAN architecture is especially well suited for texture synthesis of arbitrary large size, a novel application of adversarial methods.

%In any case, we exploit the convolutional architecture to achieve good large scale properties. 
%Good overview of texture synthesis and analysis: 
%mention limits in texture size

In the experiments in Section~\ref{sec:experiments} we will examine these points in detail.
In the next section we describe the spatial GAN architecture.

\section{The SGAN Model}
\label{sec:model}
\begin{figure}
\center{
\subfigure{\includegraphics[width=0.8\textwidth]{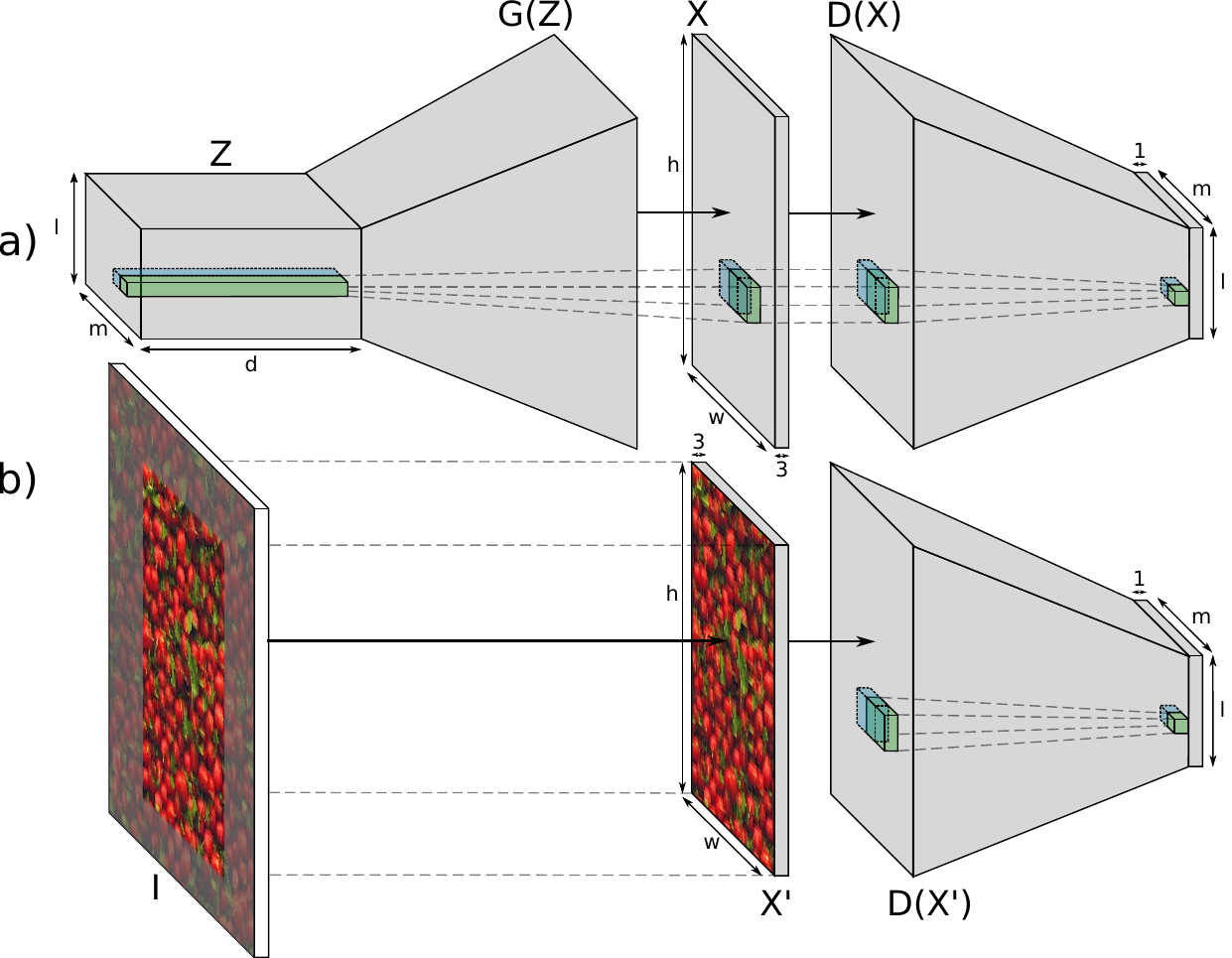} }
}
\caption{Spatial GAN model (SGAN): a generator $G$ transforms a spatial noise array $Z \in \mathbb{R}^{l \times m \times d}$  into an RGB image $X \in \mathbb{R}^{h \times w \times 3}$ via a stack of fractionally strided convolution layers. The discriminator $D$ is fed either a generated image ($X$ - case (a)) or a rectangular patch extracted from an image $I$ of a database ($X'$ - case (b)). It uses a stack of convolutional layers to output a 2D field of probabilities for fake/real ($X$ vs. $X'$) data. The detailed architecture of both generator and discriminator (the ``funnels'') is akin to~\cite{RadfordMC15}, but varies in having exclusively convolutional layers and a potentially different number of hidden layers. The subnetwork that projects a single vector of the array $Z$, i.e. $\bm{z}_{\lambda\mu} \in \mathbb{R}^d$, to the generated output image is equivalent to a standard GAN (see e.g. the green blocks in the figure). However, non-overlapping vectors have overlapping projective fields in the output. In this view, SGAN is a convolutional roll-out of GANs.}
\label{fig:SGAN}
\end{figure}
The key idea behind Generative Adversarial Networks~\cite{Goodfellow14} is to simultaneously learn a generator network $G$ and a discriminator network $D$. The task of $G(\bm{z})$ is to map a randomly sampled vector $\bm{z} \in \mathbb{R}^d$ from a prior distribution $p_{\bm z}(\bm z)$ to a sample $X \in \mathbb{R}^{h \times w \times 3}$ in the image data space. The discriminator $D(X)$ outputs a scalar representing the probability that $X$ is from real training data and not from the generator $G$. Learning is motivated from game theory: the generator $G$ tries to fool the discriminator into classifying generated data as real one, while the discriminator tries to discriminate real from generated data. As both $G$ and $D$ adapt over time, $G$ generates data that gets close to the input data distribution.

%\footnote{Potential TODO: describe min-max training and formula? Maybe a bit more explanation about the motivation for GANs?}

The SGAN generalizes the generator $G(Z)$ to map a tensor $Z \in \mathbb{R}^{l \times m \times d}$  to an image $X \in \mathbb{R}^{h \times w \times 3}$, see Figure~\ref{fig:SGAN}. We call $l$ and $m$ the spatial dimensions and $d$ the number of channels. Like in GANs, $Z$ is sampled from a (simple) prior distribution: $Z \sim p_Z(Z)$. We restricted our experiments to having each slice of $Z$ at position $\lambda$ and $\mu$, i.e. $\bm{z}_{\lambda\mu} \in \mathbb{R}^d$, independently sampled from $p_{\bm z}(\bm z)$, where $\lambda, \mu \in \mathbb{N}$ with $1\leq \lambda \leq l$ and $1\leq \mu \leq m$. 
Note that the architecture of the GAN puts a constraint on the dimensions $l,m,h,w$ -- if we have a network with $k$ convolution layers with stride $\frac{1}{2}$ and same zero padding (see e.g.~\cite{Dumoulin2016}) then $\frac{h}{l}=\frac{w}{m}=2^k$.
% TODO: check that  zero padding is spelled identical throughout the paper!

%We denote the count of pixels equal to one spatial dimension of $Z$ as $r=\frac{H}{U}$.
%Similar to the extension of the generator, 

Similarly to the way we extended the generator, the discriminator $D$ maps to a two-dimensional field of size $l \times m$ containing probabilities that indicate if an input $X$ (or $X'$) is real or generated. 
In order to apply the SGAN to a target texture $I$, we need to use $I$ to define the true data distribution $p_{\text{data}}$. To this end we extract rectangular patches $X'\in \mathbb{R}^{h \times w \times 3}$ from the image $I$ at random positions.
% which then and input it to the network $D$.
%In theory, we are flexible w.r.t. the size of $X'$ and its relation to $I$, e.g. we could feed the whole input image $I=X'$ to the discriminator $D$. %Alternatively, to speed up training, one may also use a minibatch approach: sample smaller rectangular patches $X'$ from random locations in $I$ and use these for the distribution $p_{\text{data}}$.
We chose $X'$ to be of the same size as the samples of the generator $X=G(Z)$ - otherwise GAN training failed in our experiments. For the same reason we chose symmetric architectures for $G$ and $D$, i.e. $D(G(Z))$ has the same spatial dimensions as $Z$.

%\improvement{we should also add that the G and D are symmetric, i.e. in particular receptive and projective fields are the same} 

%\improvement{REMOVE THIS?: This may be due to the fact that $D$ needs to handle samples $G(Z)$ influenced by the edges of the array.  In order to keep the distribution of $X' \sim p_{\text{data}}(X)$ with the same properties as $G(Z), Z \sim p(Z)$, we need to sample a similar proportion of such edge-influenced images from the training image distribution $p_{\text{data}}$. This is achieved in a straightforward way if we use equal sizes $H \times W$ pixels for images $X$ and $X'$, which ensures that edge patches are also learned well.}

%, including some with a distribution 

%TODO: explain it works best when $H=H',W=W'$
%we can take as one extreme the whole training image as discriminator input, or we can take small minibatches from it created in a sliding window manner
%

Both the generator $G(Z)$ and the discriminator $D(X)$ are derived from the architecture of~\cite{RadfordMC15}. In contrast to the original architecture however, spatial GANs forgo any fully connected layers - the networks are purely convolutional. This allows for manipulation of the spatial dimensions (i.e. $l$ and $m$) without any changes in the weights. Hence, a network trained to generate small images is able to generate a much larger image during deployment, which matches the local statistics of the training data. 

%Also, the convolutional architecture allows different image sizes for G and for the data in the value function - we write $L' \times M'$ for training images and $L \times M$ for generated images during learning. TODO remove the L' and M', only L,M

We optimize the discriminator (and the generator) simultaneously over all spatial dimensions:
%:, respectively $L'$ and $M'$ 

\begin{align}
 \min_G \max_D V(D,G) =& \frac{1}{lm} \sum_{\lambda=1}^l \sum_{\mu=1}^m \mathbb{E}_{Z \sim p_Z(Z)} \left[ \log \left( 1 - D_{\lambda\mu}(G(Z)) \right) \right]  \nonumber \\
  +& \frac{1}{lm} \sum_{\lambda=1}^{l} \sum_{\mu=1}^{m} \mathbb{E}_{X' \sim p_{\mathrm{data}}(X)} \left[ \log D_{\lambda\mu}(X') \right]
  \label{eq:sgan_vfunc}
\end{align}

In this formula the first row corresponds to Figure~\ref{fig:SGAN}(a), and the second row to Figure~\ref{fig:SGAN}(b). In practice, it is helpful to apply the trick of ~\cite{Goodfellow14} and minimize $- \log(D(G(Z)))$ instead of $\log(1-D(G(Z)))$.  

Note that the model describes a stochastic process over the image space. In particular, as the generator $G$ is purely convolutional and each $\bm{z}_{\lambda\mu}$ is identically distributed and independent of its location $\lambda$ and $\mu$, the generated data is translation-invariant. Hence the process is \textit{stationary} with respect to translations.

To show that the process is also strong mixing, we first need to define the projective field (PF) of a spatial patch $\hat Z$ of $Z$ as the smallest patch $\hat X$ of the image $X$ which contains all affected pixels of $X=G(Z)$ under all possible changes of $\hat Z$. In full analogy, we refer to the receptive field (RF) of a patch in $D(X)$ as the corresponding minimal patch in $X$ which affects it.
Assume then two non-overlapping patches from the generated data, $A$ and $B$. Additionally, take their respective projective fields $Z_A, Z_B$ to be non-overlapping - this can be always achieved as projective fields are finite, but the array $Z$ can be made arbitrarily large. The generated data in $A$ and $B$ is then independently generated. The process is hence \textit{strong mixing} (with the length scale of the projective field), which implies it is also \textit{ergodic}.

 %(i.e. the smallest possible subsets of $Z$ that determine the data in $A$ or $B$ under $G$)
\section{Experiments}
\label{sec:experiments}

\subsection{Architectural details and speed}

For the following experiments, we used an architecture close to the DCGAN setup~\cite{RadfordMC15}: convolutional layers with stride $\frac{1}{2}$ in the generator, convolutional layers with stride 2 in the discriminator, kernels of size 5x5 and zero padding. 
We used a uniform distribution for $p_{\bm z}(\bm z)$ with support in $[-1,1]$.
%This means that a spatial dimension of $Z$ results to 32 output pixels, and the effective receptive field size is 125, see~\cite{Dumoulin2016} for details. 
%We fed sample images to the SGAN of size $N=M=128$, and use the same size for the generated samples.
%When doing experiments with 6 layers instead, we increased the image size to 256 pixels.
%We used 64 filters in the first conv. layer and doubled them every additional layer.
Depending on the size and structure of the texture to be learned, we used networks of different complexity, see Table \ref{archi}. We used networks with identical depths in $G$ and $D$. The sizes of filter banks of $D$ were chosen to be in reverse to those of $G$, yielding more channels for smaller spatial representations.
We denote with SGAN\textit{x} that we have \textit{x} layers depth in $G$ and $D$.
We applied batch normalization on all layers, except the output layer of $G$, the input and output layers of $D$.
All network weights were initialized as 0-mean Gaussians with $\sigma=0.02$.

%, and same size for the images $X$ and $X'$ fed to $G$ and $D$.

We tried different sizes for the image patches $X$. Note that the spatial dimensions of $Z$ and $X$ are dependent, $h=rl$ and $w=rm$.
Both setting $h=w=640$ or $l=m=4$ and adjusting for the respective depending variables worked similarly well, despite different relative impact of the zero padded boundaries.

%TODO tradeoff if 128px patch then 32 sample mini, if 640 resolution than 2 or even 1 in minibatch, a tradeoff

Our code was implemented in Theano and tested on an Nvidia Tesla K80 GPU.
The texture generation speeds of a trained SGAN with different architectures and image sizes are shown in Table~\ref{speeds}. Generation with $G$ is very fast, as is expected for a single forward pass. Forward pass generation in TextureNet~\cite{ulyanov16texture} is significantly slower (20ms, according to their publication) than SGAN (5ms) for the 256 pixel resolution, despite the fact that TextureNet uses fewer filters (8 to 40 channels per convolution layer) than we do (64 to 512 filters). The simpler SGAN architecture avoids the multiple scales and join operations of TextureNet, rendering it more computationally efficient.
As expected, the method of Gatys~\cite{Gatys2015b} is orders of magnitude slower due to the iterative optimization required.
%TODO mention they have a multi scale approach and concatenations, while we have noise only at the bottom layer.  

There are initial time costs for training the SGAN on the target textures. For optimization we used ADAM~\cite{KingmaB14} with parameters as in~\cite{RadfordMC15} and 32 samples per minibatch. For the simple textures from Section~\ref{sec_smalli}, subjective assessment indicates that generated images are close to their final quality after roughly 10 minutes of training. The more complex textures of Sections \ref{sec_largei} required around 30 minutes of training. 

Training times TextureNet~\cite{ulyanov16texture} requires a few hours per texture. We could not compare this directly with SGAN on the same machine and textures, but we assume that SGAN trains more efficiently than TextureNet because of its simpler architecture.
% - lacking multi scales and the large pretrained VGG19 descriptor network.

The exact time and number of iterations required for training SGANs depends on the structure of the target texture. A general problem in GAN training is that it is often required to monitor the results and stop training -- as occasionally overtraining may lead to degeneracy or image quality degradation.
% (e.g. contrast corruption occurs sometimes).

%GAN degeneracy can be always an issue.
%Contrast corruption occurs sometimes, can be manually corrected but %still would be better if it can be fixed. For example .

\begin{table}[t]
\centering
\begin{tabular}{|c|c|c|c|}
\hline 
 & SGAN4 & SGAN5 & SGAN6 \\ 
\hline 
$Z$ channel number $d$ & 20 & 50 & 100 \\ 
\hline 
$r=h/l=w/m=$ & 16 & 32 & 64 \\ 
\hline 
PF/RF size & 61 & 125 & 253 \\ 
\hline 
generator $G$ filters &  256-128-64 & 512-256-128-64 & 1024-512-256-128-64  \\ 
\hline 
discriminator $D$ filters & 64-128-256 & 64-128-256-512 & 64-128-256-512-1024  \\ 
\hline 
\end{tabular} 
\caption{Details of the SGAN architecture with 4,5 or 6 layers. The calculation of the projective and receptive field (PF and RF) sizes is examined in detail in Appendix I.}\label{archi}
\end{table}

%\hline $U$ & 4 & 4 & 4 \\ 

\begin{table}[t]
\centering
\begin{tabular}{|c|c|c|c|c|}
\hline
 & SGAN4 & SGAN5 & TextureNet & Gatys  \\ 
\hline 
256x256 px & .005s & .006s & 0.020s & 10s \\ 
\hline 
512x512 px & .013s & .019s & - & -\\ 
\hline 
1024x1024 px & .047s & .07s & -& -\\ 
\hline 
2048x2048 px & .178s & .269s& -& -\\ 
\hline 
\end{tabular} 
\caption{Time required for generating an output texture of certain size. The SGAN architecture is faster than both TextureNet~\cite{ulyanov16texture} and Gatys~\cite{Gatys2015b}. The time costs per calculated pixel scale sublinearly.}\label{speeds}
\end{table}

\subsection{Examples of generated textures}
\subsubsection{Single small image}\label{sec_smalli}
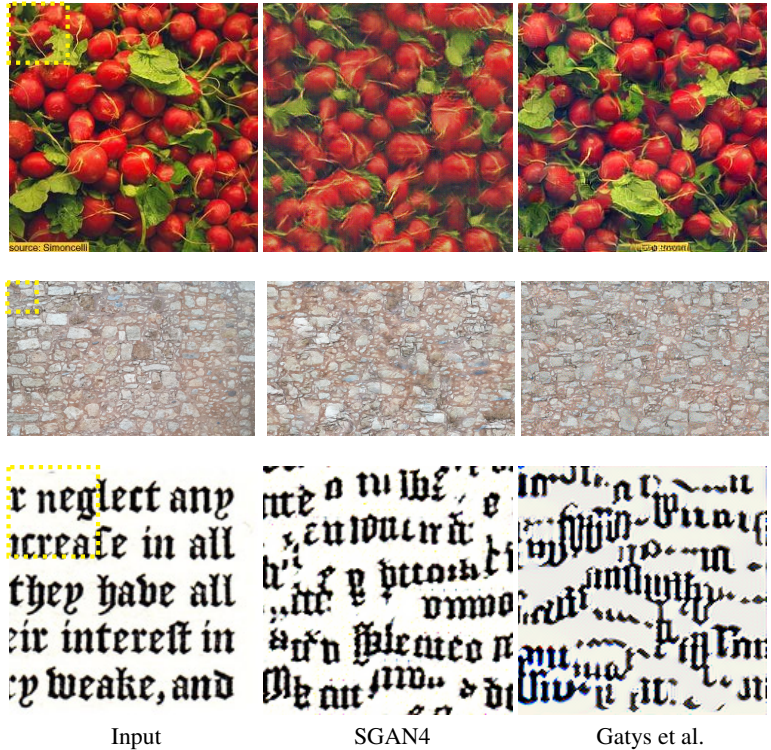
\begin{figure}
\centering
\subfigure{\begin{tikzpicture}
    \node[anchor=south west,inner sep=0] at (0,0) {\includegraphics[height=3.3cm]{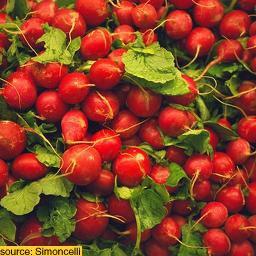}};
    \draw[yellow,dotted,ultra thick] (0,3.3) rectangle (0.786,3.3-0.786);
\end{tikzpicture} }%\includegraphics[height=3.3cm]{img/texture/simoncelli_source.jpg}
\subfigure{\includegraphics[height=3.3cm]{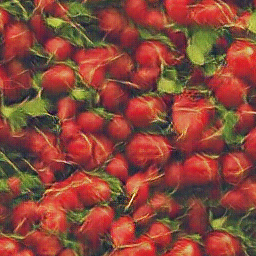}}
\subfigure{\includegraphics[height=3.3cm]{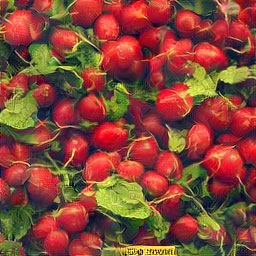} }

\subfigure{\begin{tikzpicture}
    \node[anchor=south west,inner sep=0] at (0,0) {\includegraphics[height=2.051cm]{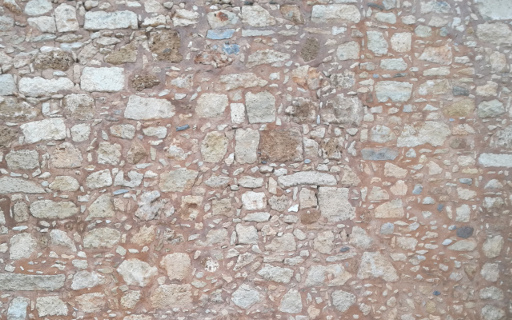}};
    \draw[yellow,dotted,ultra thick] (0,2.051) rectangle (0.39,2.051-0.39);
\end{tikzpicture} }
\subfigure{\includegraphics[height=2.051cm]{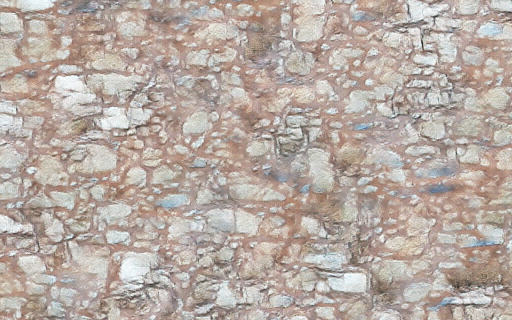}}
\subfigure{\includegraphics[height=2.051cm]{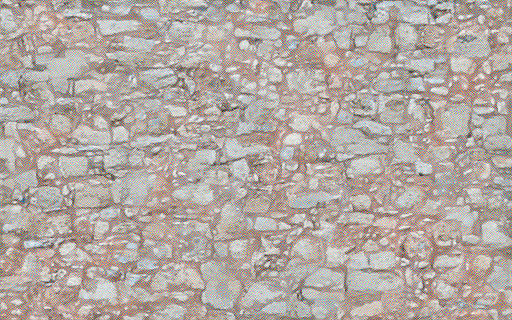} }

%stones_training_data.jpg textureGAN_TEXstones_scaled_z2040_cr.jpg GATYSstones.jpg
\subfigure[Input]{\begin{tikzpicture}
    \node[anchor=south west,inner sep=0] at (0,0) {\includegraphics[trim={392px 0 0 0},clip,height=3.3cm]{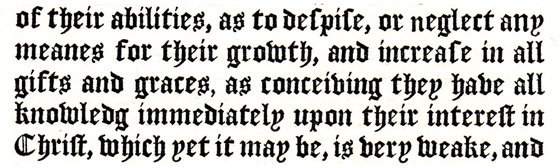}};
    \draw[yellow,dotted,ultra thick] (0,3.3) rectangle (01.19,3.3-1.19);
\end{tikzpicture}}
\subfigure[SGAN4]{\includegraphics[trim={392px 0 0 0},clip,height=3.3cm]{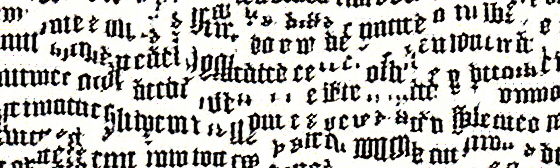}}
\subfigure[Gatys et al.]{\includegraphics[trim={392px 0 0 0},clip,height=3.3cm]{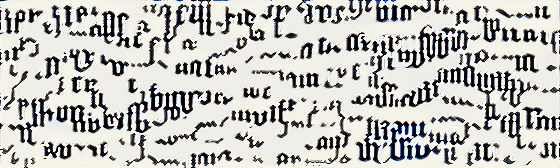} }

%\subfigure{\includegraphics[height=3.3cm]{img/textureGAN_TEXefros_4L_z10_ngf32_cr.png} }
%\subfigure[Image]{\includegraphics[height=3.3cm]{img/bible_crop_cr.jpg} }
%\subfigure[Gatys]{\includegraphics[height=3.3cm]{img/bible_crop-bible_crop-vgg19.jpg} }
%\subfigure[GAN texture synthesis]{\includegraphics[height=3.3cm]{img/textureGAN_BIBLE2_TEX40_crcr.png} }
%\subfigure[seamless large texture]{\includegraphics[height=3.3cm]{img/textureGAN_BIBLE2_TEX40_cr.png} }
\caption{Texture synthesis of small texture images of different size - the results of SGAN are comparable to Gatys~\cite{Gatys2015b}. In the top left corner of the images we indicate the receptive field of size 61 pixels of SGAN4 to visually illustrate their size relative to the texture image sizes. \label{imgtexture}}
\end{figure}

A common way to measure quality of texture synthesis is to visually evaluate the generated samples. A setup with small input textures allows a direct comparison between SGAN and the method of Gatys~\cite{Gatys2015b}.
For these examples, we used an SGAN with 4 layers. Figure \ref{imgtexture} shows our results for textures coming from a stationary and ergodic process. The textures of radishes and stones (top rows) are also mixing, while the letters (bottom row) are mixing horizontally, but not vertically. The texture synthesis of both SGAN and Gatys fail to preserve the row regularity of the source image.
% and that is why the resulting letters do not have straight rows. 

% This meant that the receptive fields were of size 61 pixels, which 
%, scale sensitivity - our method adapts better to that issue%
%Gatys also cannot learn from thw whole image if hererogeneous scales, or ractangular
%also cannot extend seamlessly to larger target image - either optimization or other issues with Gatys...
%also for small texture we need no more than 10 minutes - comparable to the costs of Gatys
%radish - from Urs
%gothic or arabic - from me, show that is fails in Gatys - actually works when we tune the parameters...%
%Gothic, Radish - can compare with Gatys
%consider simplifying thism only 3 columns, without the indcividual patch ? make correct size by cropping the GAN appropriately

%256 pixels in height 3.3cm, box 61 pixels wide -> 61/256*3.3=0.786
% 320 pixels -> 61/320*2.051=0.39
% 168 pixels height ->  61/168*3.3=1.19
 
\subsubsection{Single large image }\label{sec_largei}
%Gaudi, Miro, Starry Night - put in large image category? next to Bruegel

%\begin{figure}
%\subfigure{\includegraphics[height=7.9cm]{img/miro.jpg} }
%\subfigure{\includegraphics[height=7.9cm]{img/textureGAN4l_MIRO_TEX15__0.png} }
%\subfigure{\includegraphics[height=7.9cm]{img/textureGAN_9frames_MIRO3_TEX20__0_contrast.png} }
%\end{figure}

%http://tex.stackexchange.com/questions/57418/crop-an-inserted-image
%trim={<left> <lower> <right> <upper>}
%1371 647 pixels - diff by 724ox
%other stuff is 2048 large - clip 1401pixels
%641 pixels images
%125/647pixels*4.5 0.87
%61 0.42
%253 0.176

%https://www.google.com/maps/@41.4033724,2.1551133,292m/data=!3m1!1e3
%The top row shows the spatial auto-correlations of each image

%Our method works exceptionally well with large textures where one has enough randomness but also specific structures of certain size. Satellite maps provide such example textures. Our sliding patch approach to input data being fed in the GAN is especially well suited to capture the information 

Satellite images provide interesting examples of macroscopic structures with texture-like properties. City blocks in particular resemble realizations from a stationary stochastic process: different city blocks have similar visual properties (color, size).
Mixing occurs on a characteristic length scale given by the major streets. 
% There is also some mixing, dependent on the length of the long streets giving structure to the city.

%TODO mention Mountains and fields do not work so well?.

%Maps of cities represent realizations of stationary processes and the regularity of city blocks

Figure \ref{imgbarca} shows that our method works better than~\cite{Gatys2015b} on satellite images, here concretely a single image of Barcelona.\footnote{\url{www.google.com/maps/@41.4033724,2.1551133,292m/data=!3m1!1e3}} SGAN creates a city-like structure, whereas Gatys' method generates less structure and detail.
We interpret this in the following way: the SGAN is trained specifically on the input image and utilizes all its model power to learn the statistics of this particular texture. 
Gatys~\cite{Gatys2015b} relies on pretrained filters learned on the ImageNet dataset, which generalize well to a large set of images (but apparently not well to satellite imagery) and fails to model salient features of this city image, in particular the street grid orientation.

%is big but cannot hold every possible image, and so apparently the city block statistics are different than the .
%cannot possible image in the world may not be perfectly suited to the specific

\begin{figure}[tb]
\centering
\subfigure[Input]{\includegraphics[height=2.2cm]{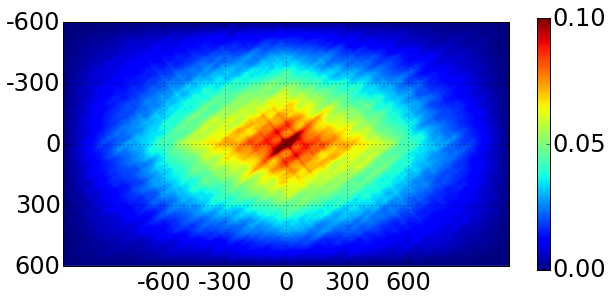} }
\subfigure[SGAN5]{\includegraphics[height=2.2cm]{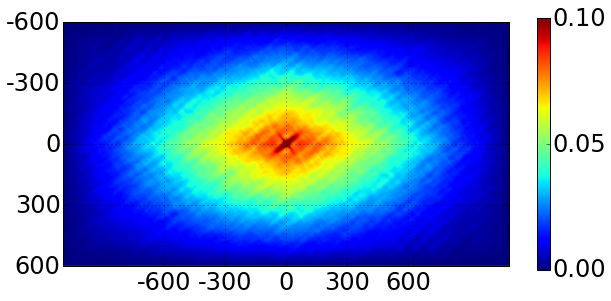} }
\subfigure[Gatys et al.]{\includegraphics[height=2.2cm]{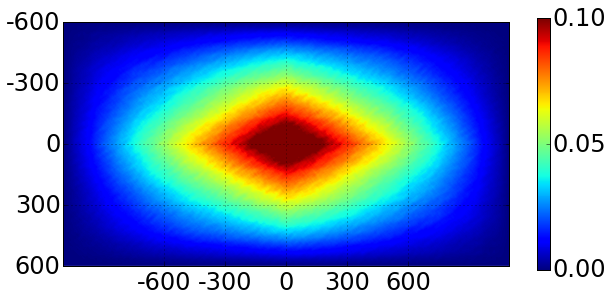} }
\caption{Spatial autocorrelation (AC) of the Barcelona city example from Figure \ref{imgbarca}: the preferred directions of the city streets are clearly visible in the centre of the AC of the input texture. The AC of the SGAN texture reflects the structure  much better than the result of Gatys et al.}\label{imgac}
\end{figure}

To indicate the superior quality of SGAN for that texture the spatial auto-correlation (AC) of the original and synthesized textures from Figure \ref{imgbarca} are shown on Figure \ref{imgac}. We calculate the AC on whole images of size 1371x647 pixels. The AC of the original and the SGAN5 generation are similar to one another and show clearly the directions of the street grid. In contrast, the AC of Gatys' texture looks more isotropic than the original, indicating loss of visually important information.

\begin{figure}
\centering
\subfigure[SGAN4 \label{barca4l}]{\begin{tikzpicture}
    \node[anchor=south west,inner sep=0] at (0,0) {\includegraphics[height=4.5cm]{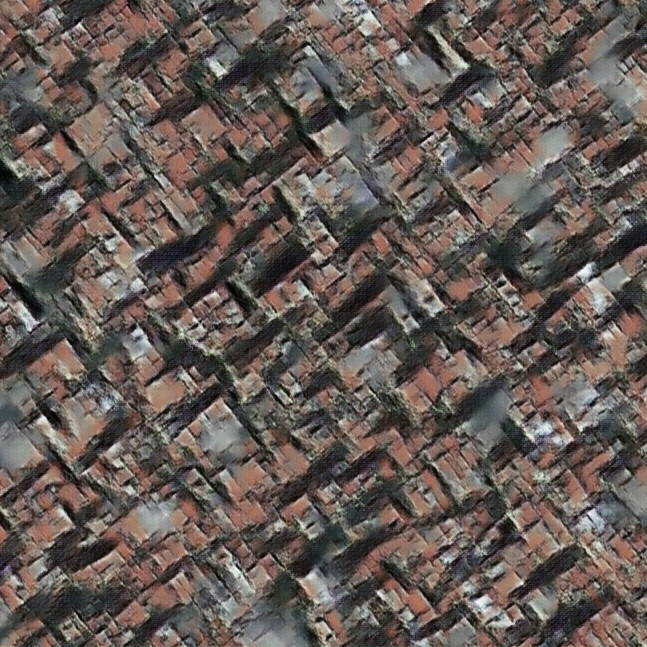}};
    \draw[yellow,dotted,ultra thick] (0,4.5) rectangle (0.42,4.5-0.42);
\end{tikzpicture} }
\subfigure[SGAN5 ]{\begin{tikzpicture}
    \node[anchor=south west,inner sep=0] at (0,0) {\includegraphics[height=4.5cm]{img/barca/textureGAN_BARCAC_TEX90__0_647.jpg} };
    \draw[yellow,dotted,ultra thick] (0,4.5) rectangle (0.87,4.5-0.87);
\end{tikzpicture}}
\subfigure[SGAN6]{ \begin{tikzpicture}
    \node[anchor=south west,inner sep=0] at (0,0) {\includegraphics[height=4.5cm]{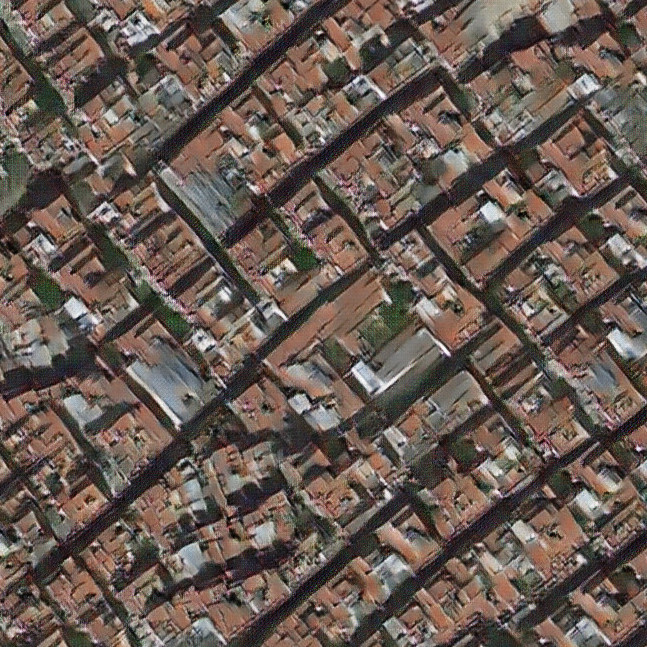}};
    \draw[yellow,dotted,ultra thick] (0,4.5) rectangle (1.76,4.5-1.76);
\end{tikzpicture}}
\caption{SGAN with $G,D$ layers of depth 4,5,6 changes the resulting output texture (cropped to size 647x647 pixels). Bigger models exhibit less mixing and generate longer streets. \label{img456}}
\end{figure}

% The texture synthesized by our method has AC much closer to the original than the one synthesized by the other method. We can see in the center of the image , unlike the AC of the other method which has a more blurry structure.
Figure \ref{img456} illustrates the effects of different network depths on the SGAN generated outputs.
%The effective receptive field size relative to the structures in the textures also makes a difference.
More layers and larger receptive fields as in SGAN6 allow larger structures to be learned and longer streets emerge, i.e., there is less mixing and more regularity at a given scale.

% in the randomly sampled textures when using larger models\improvement{mention less mixing for SGAN6}.

%\subfigure[Image]{\includegraphics[height=3.3cm]{img/antoni-gaudi-ceramic-mosaic-design-guell-park.jpg} }
%\subfigure[GAN texture synthesis]{\includegraphics[height=3.3cm]{img/GAUDI_TEX8cr.png} }
%\subfigure[seamless large texture]{\includegraphics[height=3.3cm]{img/textureGAN4l_GAUDI_TEX10__0_cr.png} }

%The Fight Between Carnival and Lent by Bruegel the Elder, painted in 1559

%\begin{figure}[tb]
%\subfigure[]{\includegraphics[height=7.9cm]%{img/hieronymus_bosch11.jpg} }
%\subfigure[5 Layers]{\includegraphics[height=7.9cm]%{img/textureGAN_GAST_TEX25_contrast.png} }
%\subfigure[4 Layers]{\includegraphics[height=7.9cm]{img/textureGAN4l_BRUEGEL192_TEX10__0_contrast.png}}
%\end{figure}
%\subfigure[GAN texture synthesis]{\includegraphics[height=3.3cm]{img/GAST_TEX25_cr.png} }

\subsubsection{Composite textures from multiple images}\label{refErgo}
\begin{figure}[tb]
\centering
%\subfigure[Image]{\includegraphics[height=3.3cm]{img/image_00055.jpg}  }
%\subfigure[GAN patches synthesis]{\includegraphics[height=3.3cm]{img/FLOWERS102_TEX249_cr.png} }

%so make 384 pixels large
%trim={664px 664 1000 1000},clip,
\subfigure[\textbf{a)} Input]{\includegraphics[height=4.5cm]{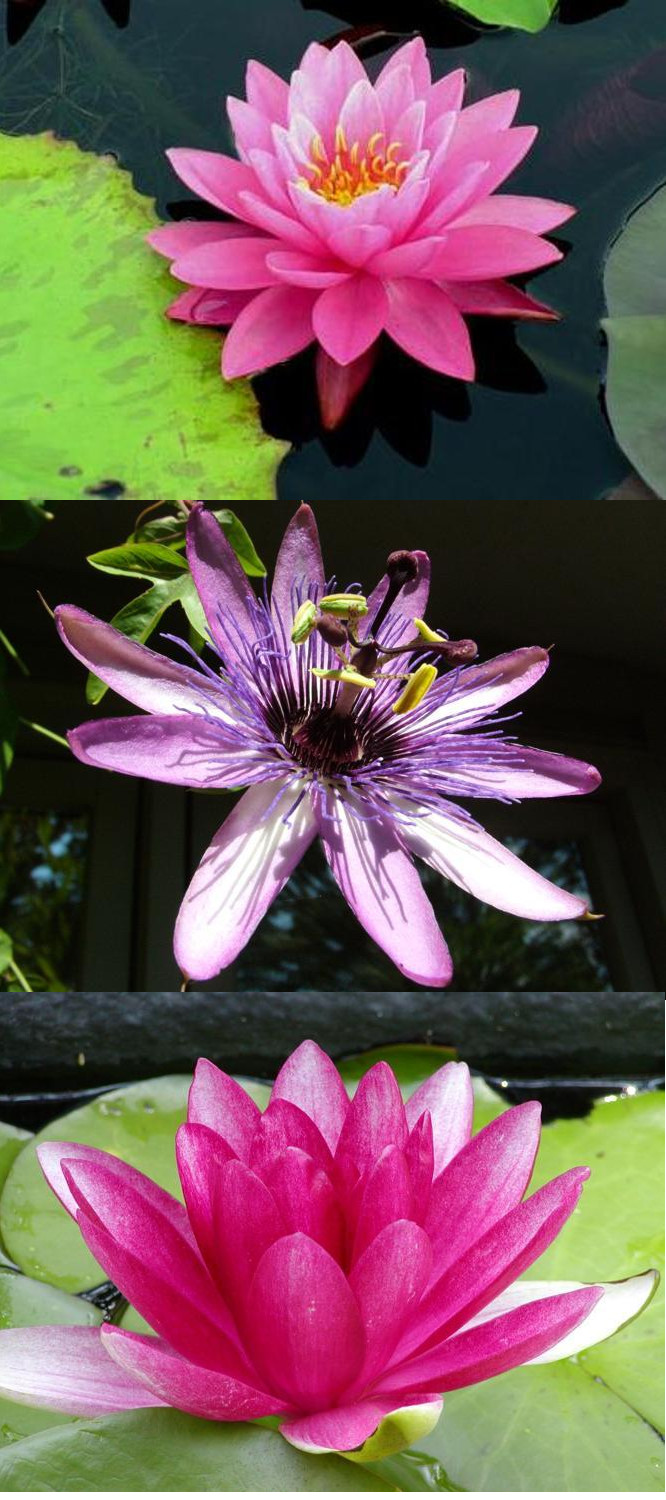} }
\subfigure[\textbf{b)} SGAN5]{\includegraphics[height=4.5cm]{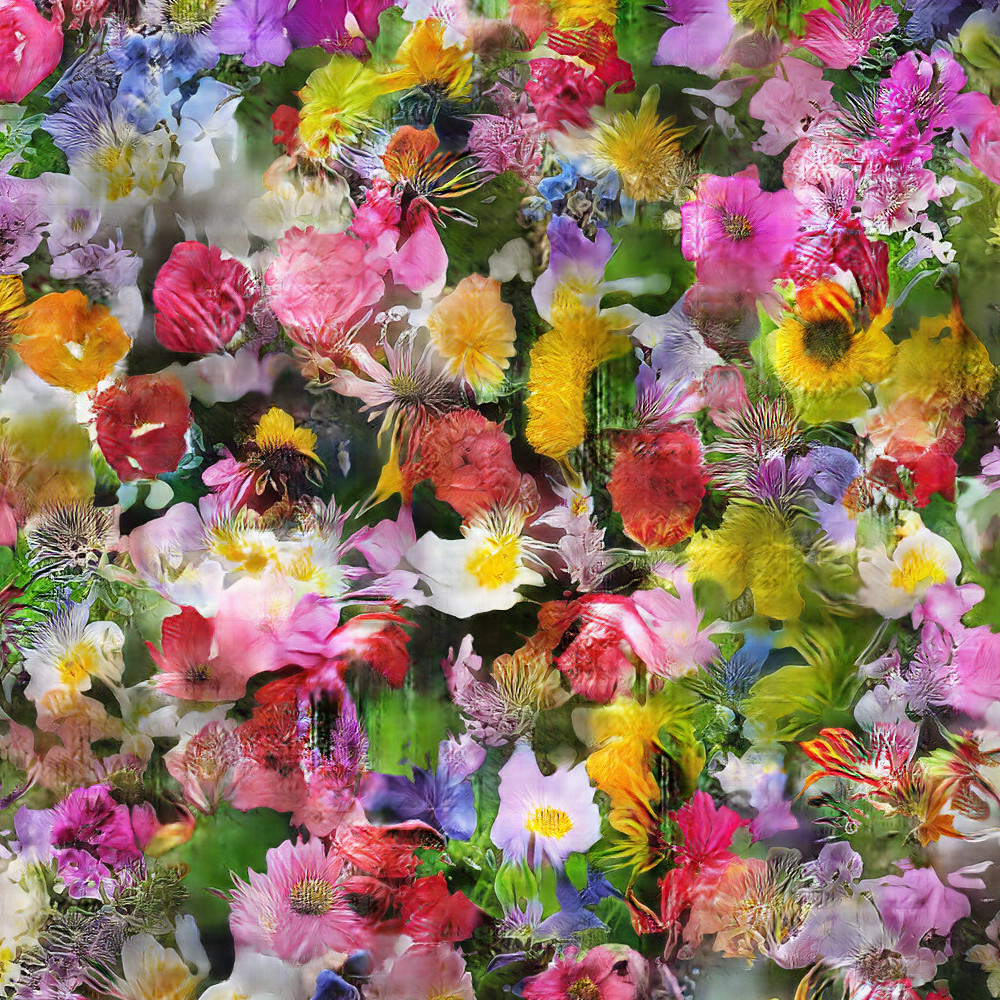} }
\subfigure[\textbf{c)} Input]{\includegraphics[height=4.5cm]{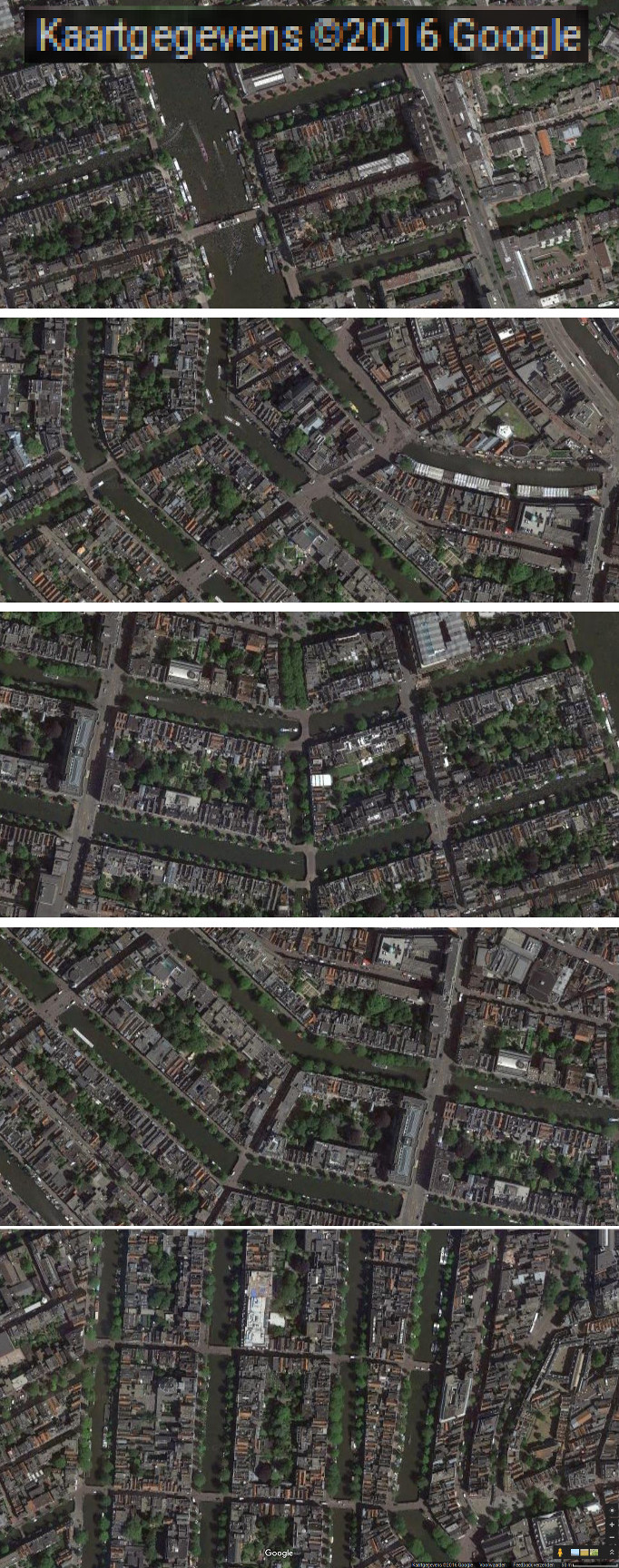} }
\subfigure[\textbf{d)} SGAN6]{\includegraphics[height=4.5cm]{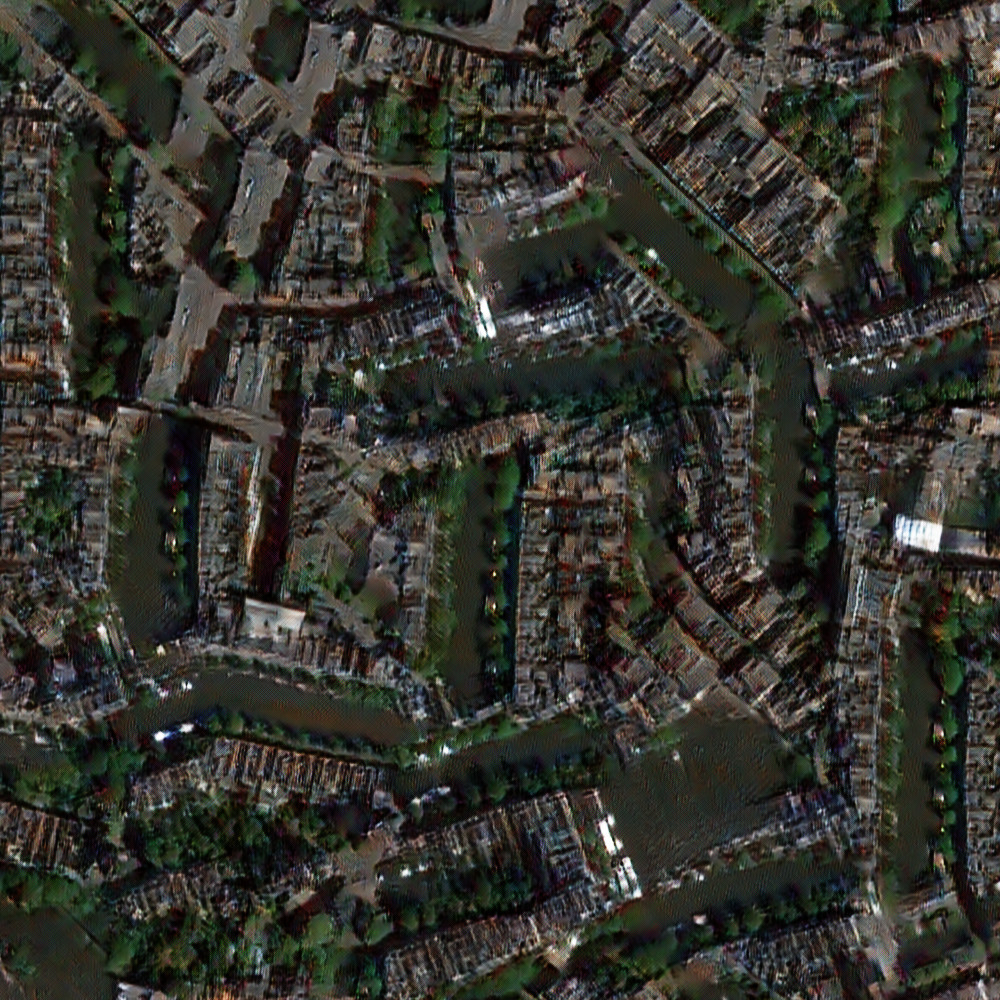} }
\caption{Learning textures from multiple images. (a) and (c) show the training data textures rescaled to fit the screen. (b) and (d) show examples of generated 1000x1000 pixel composite textures.}
\label{fusion}
\end{figure}

The GAN approach to texture synthesis can combine multiple example texture images in a natural way.
%: sample $X'$ randomly from all input textures in equation~\ref{eq:sgan_vfunc}. %This can lead to complicated textures without analog in other texture synthesis methods.
We experimented with the flowers dataset\footnote{\url{www.robots.ox.ac.uk/~vgg/data/flowers/}} containing 8189 images of various flowers, see Figure \ref{fusion} (a) for examples. We resized each image to 160 pixels in the $h$-dimension, while rescaling the $w$-dimension to preserve the aspect ratio of the original image. Then we trained an SGAN with 5 layers; each minibatch for training contained 128x128 pixel patches extracted at random positions from randomly selected input flower images. 
This is an example of a non-ergodic stochastic process since the input images are quite different from one another.

%: images are different from one another and the sample averages of the patches of one image would not converge to the ensemble averages of the patches over all images.
% Still, this is a stationary process since
A sample from the generated composite texture is shown in Figure \ref{fusion} (b).
The algorithm generates a variety of natural looking flowers but cannot blend them smoothly since it was trained on a dataset of single flower images.
Still, the final result looks aesthetic and such fusion of large image datasets for texture learning has great potential for photo mosaic applications.

%it shows locally the appearance of different flowers, but cannot blend them smoothly since it has only seen single flowers in the dataset.

% SGAN can still learn from the locally consistent image patches and generate texture exhibiting characteristics of different input texture in different regions of the output image.

%TODO decide on this, or Amsterdam 5 map slices combo, or tomatoes! mention weak/middle/strong ergodicity, view images as realizations of the stochastic process.

%We also showed that we can learn textures from multiple input images.
%Effectively, we can learn locally consistent patches from multiple images 

%we can decompose images into parts and recombine them, a biologically plausible process.

In another experiment we learned a texture representing the 5 satellite images shown on Figure \ref{fusion} (c). 
The input images depict areas in the Old City of Amsterdam with different prevailing orientations.
Figure \ref{fusion} (d) demonstrates how the generated texture has orientations from all 5 inputs. Although we did not use any data augmentation for training, the angled segments join smoothly and the model learns spatial transitions between the different input textures. Overall, the Amsterdam city segments come from a more ergodic process than the flowers example, but less ergodic than the Barcelona example. 

GAN based methods can fuse several images naturally, as they are generative models that capture the statistics of the image patches they're trained with.
%Information fusion is something that GAN methods can do naturally, since they learns directly the generator and discriminator to represent the statistics on any distribution of images or local image patches.
In contrast, methods with specified image statistical descriptors~\cite{Portilla:2000,Gatys2015b} generate textures that match a single target image closely in these descriptors. Extending these methods to several images is not straight-forward.

%, but a simple solution would be to use the average descriptors of all input images. However, this is not appropriate when the target image set is big and varied, as then averaging can destroy information by collapsing multi-modal peaks in the data distribution.
% if the average has properties different than the single examples.

%In contrast, a generative model like SGAN can easily learn a more complex model of the "right" statistics without any averaging.
%Parametric  have difficulties to combine different input images in composite textures. The reason is that their approach requires calculation of the descriptor of the single input texture image, or by extension the average descriptor of the multiple input textures. In general, such 

%, even patches coming from different images.

%the SGAN samples from all input patches during training and can generate outputs exhibiting locally the properties of different inputs.
  
%flowers are 2048
%Amsterdam is 1920 

%interpolation in z space - first set all $v \times v$ dimensions the same and interpolate as in the first GAN
%learn the $G^{-1}(X)$ inverse mapping to have more control

\subsection{Extension using properties of the spatial GAN} \label{sec_tilesplit}
The spatial dimensions of $Z$ are locally independent -- output image pixels depend only on a subset of the input noise tensor $Z$.
This property of the SGAN allows two practical tricks for creation of output textures with special properties. Below we illustrate these tricks briefly, see Appendix I for details.

%or the concrete example of an architecture with 5x5 kernels and $k$ convolutional layers with stride $\frac{1}{2}$, 

%We defined already $r$ as the count of pixels equal to 1 spatial dimension of $Z$.
\subsubsection{Seamless textures} \label{sec_tile}
\begin{figure}[tb]
\centering
\includegraphics[height=4.5cm]{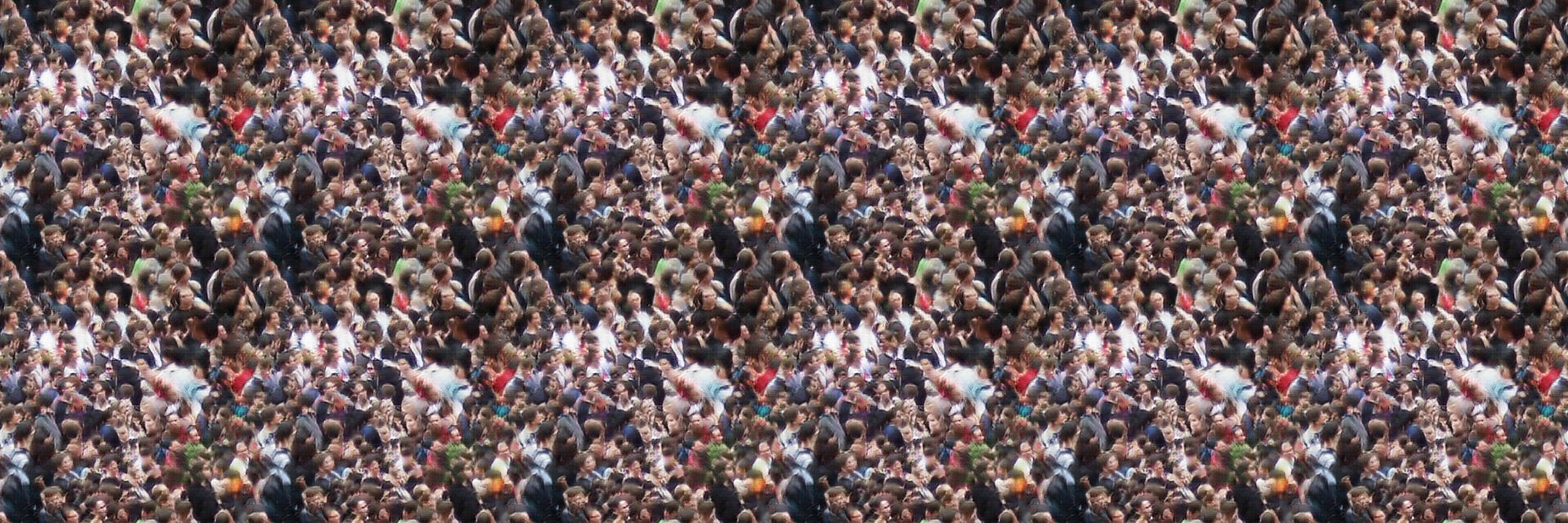} 
\caption{A 320x320 pixels texture, tiled 2 times vertically and 6 times horizontally. It is straight-forward to create seamless textures in the SGAN framework by enforcing periodic boundary conditions on $Z$.}
\label{img_tile}
\end{figure}

Seamless textures are important in computer graphics, because arbitrarily large surfaces can be covered by them. 
Suppose we want to synthesize a seamless texture of desired size $\hat{h},\hat{w}$ and generating it would require $\hat{l},\hat{m}$ spatial dimensions of $Z$ for a given SGAN model. 
Let $r=\hat{h}/\hat{l}=\hat{w}/\hat{m}$ be the ratio of the spatial dimensions of $G(Z)$ to $Z$. For notation we use Python slicing notation, where $Z_{[-i,4:]}$ indicates $i$ indices before the end of the array in the 1st dimension, the `4:' indicates all elements but the first 4 along the second dimension, and all elements along the last, not explicitly indexed, dimension.
We should sample a slightly bigger noise tensor $Z \in \mathbb{R}^{(\hat{l}+ 4) \times (\hat{m} + 4) \times d}$ and set its edges to repeat: $Z_{[:,-4:]} = Z_{[:,:4]}$ and $Z_{[-4:,:]} = Z_{[:4,:]}$. Then we can calculate $G(Z)$ and crop $2r$ pixels from each border, resulting in an image $I=G(Z)_{[2r:-2r,2r:-2r]}$ of size $\hat{h},\hat{w}$ that can be tiled in a rectangular grid as shown on Figure \ref{img_tile}.

\subsubsection{Memory efficient generation}\label{sec_split}
In addition, we can use the SGAN generator to create textures in a memory efficient way by splitting the calculation of $I=G(Z)$ into independent chunks that use less memory. This approach allows for straightforward parallelization. A potential application would be real-time 3D engines, where the SGANs could produce the currently visible parts of arbitrary large textures.

Suppose again that we have $Z \in \mathbb{R}^{\hat{l} \times \hat{m} \times d}$.
Let us split $Z$ in two along dimension $\hat{l}$. We can call the generator twice, producing $I^1=G(Z_{[:\hat{l}/2+2]})$ and $I^{2}=G(Z_{[\hat{l}/2-2:]})$, with each call using approximately half the memory than a call to the whole $G(Z)$. To create the desired large output, we concatenate the two partially generated images, cropping their edges: 
$I = [I^1_{[:-2r]},I^{2}_{[2r:]}]$. With this approach, the only limitation is the number of pixels that can be stored, while the memory footprint in the GPU is constant. Appendix I has precise analysis of the procedure.

%, the memory footprint in the GPU of the generator call plays no role.

%"img/barca/TILE_BARCAC_TEX90__(320, 320)_(2, 6)"

%make unendingly large texture by analyzing which z dimensions contribute to which pixels
%use periodicity in the conv filters to get self-tiling textures; Easier: just use periodic Zs.

\section{Discussion}
%\subsection{Comparison between SGAN and other parametric methods for texture synthesis}
%We can discuss here what is the same and what is different to the other methods we examined in the introduction.

Our SGAN synthesizes textures by learning to generate locally consistent image patches, thus making use of the repeating structures present in most textures. The mixing length scale of the generation depends on 
% -- removed this as the PF depends on it anyways: "the network architecture and "
the projective field sizes. Choosing the best architecture depends on the specific texture image. This holds also for the algorithm of Gatys et al.~\cite{Gatys2015b}, where the parametric expressiveness of the model depends on the set of network layers used for the descriptive statistics calculation.

%The adversarial framework learns which statistics are important for an example texture $I$. GANs do not use any handcrafted features or other priors to specify what properties the desired outputs should have. The discriminator $D$ learns important statistics only from information contained in $I$. 
Rather than using handcrafted features or other priors to specify the desired properties of the output, the adversarial framework learns the relevant statistics only from information contained in the training texture.
In contrast, parametric methods specify statistical descriptors a priori, before seeing the image $I$. This includes models using the wavelet transform~\cite{Portilla:2000} or the properties of the filters of a neural network trained on a large image dataset~\cite{Gatys2015b}. Such models can generalize to many textures, but are not universal -- sometimes it is better to train a generative model for a single texture, as our examples from Section~\ref{sec_largei} show.

\cite{ulyanov16texture} is an interesting case because it describes a generative model that takes as inputs noise vectors and produces images with desired statistics. This is analogous to the generator $G$ in the GAN. However, features extracted from pre-trained discriminative networks (as in~\cite{Gatys2015b}) play the role of a discriminator function, in contrast to learned discriminators in adversarial frameworks.
%less data driven and more model assumption based 
%However, they do not explore how they scale to large images, much larger than what they are trained on.

%In terms of performance, our SGAN is similar to the method of~\cite{ulyanov16texture}: generation is very fast in a forward pass, however this comes at the cost of having to train initially a neural network for every example texture.

%\subsubsection{Good and bad textures}
The examples in Sections \ref{sec_smalli} and \ref{sec_largei} show that our method deals well with texture images, corresponding to realizations of stationary ergodic stochastic processes that mix.
%Learning from multiple example textures also works, albeit with varying image quality and blending properties, as the examples from section \ref{refErgo} show.
% with weak ergodicity (see~\cite{DCC2013} for a formal treatment)
%The stationarity property of textures is necessary for our method.
%Intuitively speaking, generated images have the same statistics independent of location.
% textures should have a repetitive character and local neighborhoods have similar statistics.
%Natural images with other properties -- e.g. 
It is not possible to learn statistical dependencies that exceed the projective field size with SGANs.

%In contrast, the method of Gatys works well with such images.
%Images without such properties cannot be learned with our SGAN, e.g. that Van Gogh painting\footnote{\url{https://en.wikipedia.org/wiki/The_Starry_Night}} has long brush stroke correlations that makes it difficult to be learned locally.
%This affinity for locally consistent and globally random patches comes from the SGAN architecture, where as we discussed 

Synthesizing regular non-mixing textures (e.g. a chess grid pattern) is problematic for some methods~\cite{LLH04}.
The SGAN cannot learn such cases - distant pixels in the output image are independent from one another because of the strong mixing property of the SGAN generation, as discussed in Section \ref{sec:model}. For example, the model learns basic letter shapes from the text image in Figure \ref{imgtexture} (bottom row), but fails to align the "letters" into globally straight rows of text. 
The same example shows that the approach of Gatys~\cite{Gatys2015b} has similar problems synthesizing regular textures. In contrast, the parametric method of~\cite{Portilla:2000} and non-parametric instance-based methods work well with regular patterns.

%shows a text texture - we learn the basic letter shapes, but cannot learn globally straight rows of text.
%because each output pixel is generated by a small subset of adjacent elements of $Z$, see Section \ref{sec_tilesplit}
%, even though it optimizes textures w.r.t. more global image statistics than the SGAN local patches.

%\section{Conclusion}
%\unsure{merge conclusion and discussion, skip first paragraph here}
%We introduced the spatial GAN architecture and explored its application to texture synthesis. We showed numerous examples where our method works quite well and also discussed some of its limitations, comparing and contrasting it with other popular texture synthesis techniques. 

To summarize the capabilities of our method:
\begin{itemize}
  \item real-time generation of high quality textures with a single forward pass
  %, unlike most previous parametric methods where the time for optimizing the right image is often significant and scales prohibitively with the desired output texture size
  %\footnote{a benchmark should be done here (probably on the ms scale) - mention that it's just a feedforward pass once trained!}
  \item generation of images of any desired size
  \item processing time requirements scale linearly with the number of output pixels
  %\footnote{well, time is a constraint :) - focus on memory limitation? or s.th. like linear in pixels/size...}
  \item combination of separate source images into complex textures
  %\item dynamic texture generation and smooth morphing
  \item seamless texture tiles
\end{itemize}

As next steps, we would like to examine modifications allowing the learning of images with longer spatial correlations and strong regularity. Conditional GAN~\cite{RadfordMC15} architectures can help with that, as well as allow for more precise control over the generated content. Conditioning can be used to train a single network for a variety of textures simultaneously, and then blend them in novel textures.
%which is now a limitation of our local patch-oriented approach.

In future work, we plan to examine further applications of the SGAN such as style transfer, mosaic rendering, image completion, in-painting and modeling 3D data. In particular, 3D data offers intriguing possibilities in combination with texture modeling -- e.g. in machine generated fashion designs. The SGAN approach can also be applied to audio data represented as time series, and we can experiment with novel methods for audio waveform generation.
%replacing the image textures with audio samples and modifying the convolutional layers to deal with such
% TODO mention some fashion application in the outlook, learning and generating clothes, quick tests for fitting garments..

%, audio application, 

\section*{Acknowledgements}
We would like to thank Christian Bracher, Sebastian Heinz and Calvin Seward for their valuable feedback on the manuscript.

%\clearpage
%\newpage

\bibliography{bibi}{}

\begin{thebibliography}{10}

\bibitem{Dumoulin2016}
Vincent Dumoulin and Francesco Visin.
\newblock A guide to convolution arithmetic for deep learning.
\newblock {\em arXiv:1603.07285}, 2016.

\bibitem{EfrosQ}
Alexei~A. Efros and William~T. Freeman.
\newblock Image quilting for texture synthesis and transfer.
\newblock In {\em Proceedings of the 28th Annual Conference on Computer
  Graphics and Interactive Techniques}, SIGGRAPH, 2001.

\bibitem{EfrosP}
Alexei~A. Efros and Thomas~K. Leung.
\newblock Texture synthesis by non-parametric sampling.
\newblock In {\em Proceedings of the International Conference on Computer
  Vision}, 1999.

\bibitem{lapgan}
Arthur~Szlam Emily~Denton, Soumith~Chintala and Rob Fergus.
\newblock Deep generative image models using a {L}aplacian pyramid of
  adversarial networks.
\newblock In {\em Advances in Neural Information Processing Systems 28}, 2015.

\bibitem{Gatys2015b}
Leon Gatys, Alexander Ecker, and Matthias Bethge.
\newblock Texture synthesis using convolutional neural networks.
\newblock In {\em Advances in Neural Information Processing Systems 28}, 2015.

\bibitem{DCC2013}
G.~Georgiadis, A.~Chiuso, and S.~Soatto.
\newblock Texture compression.
\newblock In {\em Data Compression Conference}, March 2013.

\bibitem{Goodfellow14}
Ian~J. Goodfellow, Jean Pouget{-}Abadie, Mehdi Mirza, Bing Xu, David
  Warde{-}Farley, Sherjil Ozair, Aaron~C. Courville, and Yoshua Bengio.
\newblock Generative adversarial nets.
\newblock In {\em Advances in Neural Information Processing Systems 27}, 2014.

\bibitem{IoffeS15}
Sergey Ioffe and Christian Szegedy.
\newblock Batch normalization: Accelerating deep network training by reducing
  internal covariate shift.
\newblock In {\em Proceedings of the 32nd International Conference on Machine
  Learning}, 2015.

\bibitem{Johnson2016Perceptual}
Justin Johnson, Alexandre Alahi, and Li~Fei-Fei.
\newblock Perceptual losses for real-time style transfer and super-resolution.
\newblock In {\em European Conference on Computer Vision}, 2016.

\bibitem{KingmaB14}
Diederik~P. Kingma and Jimmy Ba.
\newblock Adam: {A} method for stochastic optimization.
\newblock {\em CoRR}, abs/1412.6980, 2014.

\bibitem{LLH04}
Yanxi Liu, Wen-Chieh Lin, and James Hays.
\newblock Near-regular texture analysis and manipulation.
\newblock In {\em ACM SIGGRAPH Papers}, 2004.

\bibitem{Portilla:2000}
Javier Portilla and Eero~P. Simoncelli.
\newblock A parametric texture model based on joint statistics of complex
  wavelet coefficients.
\newblock {\em Int. J. Comput. Vision}, 40(1), October 2000.

\bibitem{RadfordMC15}
Alec Radford, Luke Metz, and Soumith Chintala.
\newblock Unsupervised representation learning with deep convolutional
  generative adversarial networks.
\newblock {\em CoRR}, abs/1511.06434, 2015.

\bibitem{DB15a}
Jost~T. Springenberg, Alexey Dosovitskiy, Thomas Brox, and Martin Riedmiller.
\newblock Striving for simplicity: The all convolutional net.
\newblock In {\em ICLR (workshop track)}, 2015.

\bibitem{Theis2015c}
Lucas Theis and Matthias Bethge.
\newblock Generative image modeling using spatial {LSTMs}.
\newblock In {\em Advances in Neural Information Processing Systems 28}, 2015.

\bibitem{ulyanov16texture}
Dmitry Ulyanov, Vadim Lebedev, Andrea Vedaldi, and Victor Lempitsky.
\newblock Texture networks: Feed-forward synthesis of textures and stylized
  images.
\newblock In {\em International Conference on Machine Learning}, 2016.

\bibitem{egst.20091063}
Li-Yi Wie, Sylvain Lefebvre, Vivek Kwatra, and Greg Turk.
\newblock {State of the Art in Example-based Texture Synthesis}.
\newblock In M.~Pauly and G.~Greiner, editors, {\em Eurographics 2009 - State
  of the Art Reports}. The Eurographics Association, 2009.

\end{thebibliography}
\bibliographystyle{plain}

\clearpage
\newpage
\section*{Appendix I}\label{refI}

We examine in detail the projective fields (PF) and which inputs from $Z$ map to which output pixels in $G(Z)$. Let $[a,b)$ indicate a range starting at index $a$ inclusive and ending at index $b$ exclusive, which we will also call left and right border. For convenience of notation, we will write only 1D indices for the square PFs, e.g. $[0,1)$ will refer to the square field $Z_{[:1,:1]}$ in python slicing notation. For simplicity, we express the formulas valid only for the architecture we usually used for SGAN, 5x5 kernels and $k$ convolutional layers with stride $\frac{1}{2}$.

We start by examining the recursive relation between input and output of a fractionally strided convolutional layer. 

\begin{theorem}
An input $[a,b)$ has as its PF an output $[a',b')$ after applying one convolutional layer. It holds that $a' = 2a-2$ and $b' = 2b+1$.
\end{theorem}
This relation holds because of the way we implement a convolutional layer with stride $\frac{1}{2}$ in Theano, just like DCGAN~\cite{RadfordMC15} does. Note that $b'-a' = 2(b-a)+3$ which is exactly relationship 13 from~\cite{Dumoulin2016} between input and output sizes of a transposed convolution.

We can rewrite the recursive relations as a function of the initial size and count of layers $k$:

\begin{theorem}
An input $[a,b)$ has as its PF an output $[a',b')$ after $k$ convolutional. layers. It holds that $a'= a2^k - 2^{k+1}+2$ and $b' = b2^k+2^{k}-1$.
\end{theorem}
In particular, we get the PF size of a single $\bm{z}_{\lambda\mu}$ for $b=a+1$ as $PF(k)=2^{k+2}-3$, which we denote in Table~\ref{archi} as PF/RF.

With these relations, we can show why we can split without any loss of information the calculation of a big array $Z$ in two smaller volumes as in Section \ref{sec_split}.
\begin{theorem}
Let $G(Z)=I$ with $Z \in \mathbb{R}^{\hat l \times \hat m \times d}$. We can define $I^1=G(Z_{[:\hat{l}/2+2]})$ and $I^2=G(Z_{[\hat{l}/2-2:]})$. Then it holds that $I=[I^1_{[:-2r]},I^{2}_{[2r:]}]$ where $r=2^k$.
\end{theorem}
Since we split only on one of the spatial dimensions, it is enough to reason only for intervals in that dimension and not the whole 2D field.
Image  $I^1_{[:-2r]}$ in the Python slicing notation is the same as image $I^1_{[:r\hat{l}/2]}$, and its rightmost pixel has index $x_r=2^{k}\hat{l}/2-1$. 
The PF of $Z_{[\hat{l}/2+2]}$ has a left border $(\hat{l}/2+2)2^k-2^{k+1}+2= 2^{k-1}\hat{l}+2 > x_r$.
% and thus  is not influenced by any spatial element of $Z_{[\hat{l}/2+2:]}$.
%This means that for the calculation of $I^1_{[:r\hat{l}/2]}$ we have used correctly a generator call on a subset of the $Z$ volume: $I^1 = G(Z_{[:\hat{l}/2+2]})$.
The calculation of the pixel $x_r$ is not influenced by any further elements from $Z$, i.e. for any $I^{1+\delta}=G(Z_{[:\hat{l}/2+2+\delta]}),\delta \geq 0 $ it holds that $I^1_{[:r\hat{l}/2]} = I^{1+\delta}_{[:r\hat{l}/2]}$. 
This would mean that $I^1_{[:r\hat{l}/2]}$ is exactly equal to $I_{[:r\hat{l}/2]}$, the left half of the desired image $I$.
By a similar argument, the left-most pixel of $I^{2}_{[2r:]}$ is %$x^l=$
not influenced by $Z_{[:\hat{l}/2-2]}$, and $I^{2}_{[2r:]}$ is equal to $I_{[r\hat{l}/2:]}$, the right half of the desired image $I$. This proves that $I=[I^1_{[:-2r]},I^{2}_{[2r:]}]$. 

This proof shows why an overlap of 2 spatial dimensions of $Z$ is sufficient for splitting. Is it also necessary? The answer is yes, since $Z_{[:\hat{l}/2+2]}$ is the smallest set required to calculate $I^1$: the PF of $Z_{[\hat{l}/2+1]}$ has left border $(\hat{l}/2+1)2^k-2^{k+1}+2= 2^{k-1}\hat{l}-2^k+2 < x_r$ and right border $(\hat{l}/2+2)2^k+2^k-1 >x_r$, so the pixel $x_r$ is inside the PF of $Z_{[\hat{l}/2+1]}$.

%the volumes in $Z$ which affect the pixels of  and $I^{2}_{[:,-r\hat{V}/2:]}$, respectively the left and right half of the original image $I$. 
%Thus the split of $I$ is without loss of information.

A similar proof can be made for the seamless, i.e. periodic, texture case from Section \ref{sec_tile}, and we sketch it here. Making the $Z$ periodic in its spacial dimensions makes the output $I$ periodic as well. As in the previous case, we need for each border an overlap of 2, hence we need to make 4 elements along each dimension to be identical, i.e. we set $Z_{[:,-4:]} = Z_{[:,:4]}$ and $Z_{[-4:,:]} = Z_{[:4,:]}$. More of the periodic structure in $Z$ is not needed as it would be redundant.
%In that case we use a $Z$ where the first 4 and last 4 elements of $Z$ are identical, 
The output image is  $I=G(Z)_{[2r:-2r,2r:-2r]}$. It is easily shown that the leftmost and rightmost pixels $I_{[0,0]}$ and $I_{[0,-1]}$ fit together as required for a seamless texture. These pixels use information from elements of $Z$ that are equal numerically, $Z_{[:4,:4]}=Z_{[:4,-4:]}$. The relative positions of the pixel $I_{[0,0]}$ inside the PF of the volume $Z_{[:4,:4]}$ is offset exactly by 1 pixel compared to the position of $I_{[0,-1]}$ inside the PF of $Z_{[:4,-4:]}$.
\end{document}